\documentclass[letterpaper, 10 pt, conference]{ieeeconf}  
\usepackage{amsmath,amssymb,amsfonts}
\usepackage{algorithm}
\usepackage{algpseudocode}
\usepackage{algorithmicx}
\usepackage{array}
\usepackage{float}
\usepackage{textcomp}
\usepackage{stfloats}
\usepackage{url}
\usepackage{verbatim}
\usepackage{graphicx}
\usepackage{tabularx}
\usepackage{subfigure}
\usepackage{scalerel}
\newcommand\orcidicon[1]{\href{https://orcid.org/#1}{\mbox{\scalerel*{
            \includegraphics{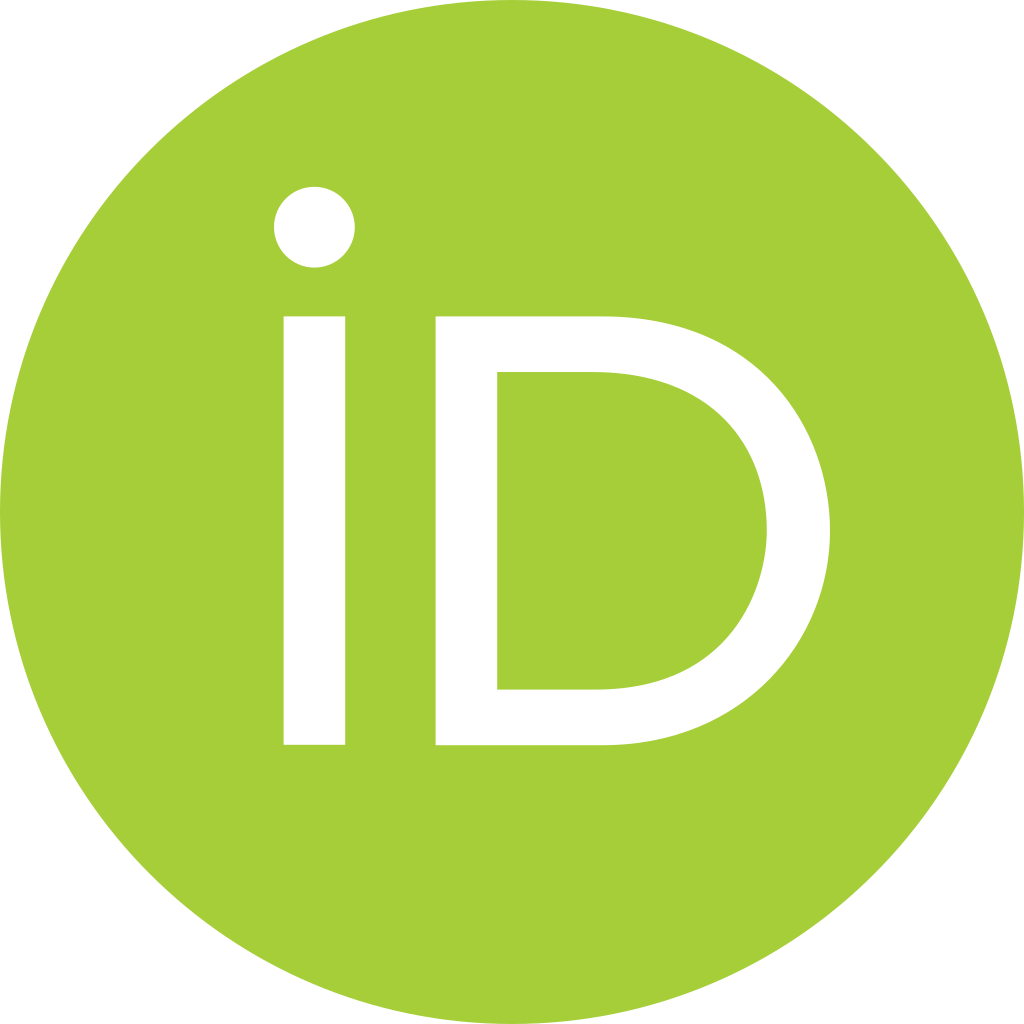}
            }{|}}}}
\usepackage{hyperref}
\usepackage{multirow}
\usepackage{xcolor}

\usepackage{filecontents}
\usepackage{cite}
\usepackage{booktabs}
\usepackage[export]{adjustbox}

\usepackage{makecell}
\usepackage{tablefootnote}

\IEEEoverridecommandlockouts                              

\title{\LARGE \bf RDMM: Fine-Tuned LLM Models for On-Device Robotic Decision Making with Enhanced Contextual Awareness in Specific Domains}

\author{Shady Nasrat$^{*\textsuperscript{\orcidicon{0000-0002-4532-7475}}}$ 
Minseong Jo$^{\textsuperscript{\orcidicon{0009-0007-3532-9016}}}$,
Myungsu Kim, 
Seonil Lee, 
Jiho Lee, 
Yeoncheol Jang, 
and Seung-joon Yi$^{\textsuperscript{\orcidicon{0000-0002-3700-4967}}}$
\thanks{*This project was funded by Police-Lab 2.0 Program(www.kipot.or.kr) funded by the Ministry of Science and ICT(MSIT, Korea) \& Korean National Police Agency(KNPA, Korea) (No. 082021D48000000) and Korea Institute for Advancement of Technology(KIAT) grant funded by the Korea Government(MOTIE)(P0008473, HRD Program for Industrial Innovation)}
\thanks{Authors are with Faculty of Electrical Engineering,
        Pusan National University, Busan, South Korea {\tt\small seungjoon.yi@pusan.ac.kr}}%
}

\begin{document}

\maketitle

\begin{abstract}
Large language models (LLMs) represent a significant advancement in integrating physical robots with AI-driven systems. We showcase the capabilities of our framework within the context of the real-world household competition. This research introduces a framework that utilizes RDMM (Robotics Decision-Making Models), which possess the capacity for decision-making within domain-specific contexts, as well as an awareness of their personal knowledge and capabilities. The framework leverages information to enhance the autonomous decision-making of the system. 
In contrast to other approaches, our focus is on real-time, on-device solutions, successfully operating on hardware with as little as 8GB of memory. Our framework incorporates visual perception models equipping robots with understanding of their environment. Additionally, the framework has integrated real-time speech recognition capabilities, thus enhancing the human-robot interaction experience.
Experimental results demonstrate that the RDMM framework can plan with an 93\% accuracy. Furthermore, we introduce a new dataset consisting of 27k planning instances, as well as 1.3k text-image annotated samples derived from the competition. The framework, benchmarks, datasets, and models developed in this work are publicly available on our GitHub repository at https://github.com/shadynasrat/RDMM.

\end{abstract}

\section{Introduction}

\begin{figure}[t] 
\centering
\setlength{\tabcolsep}{0pt} 
\includegraphics[width=\columnwidth]{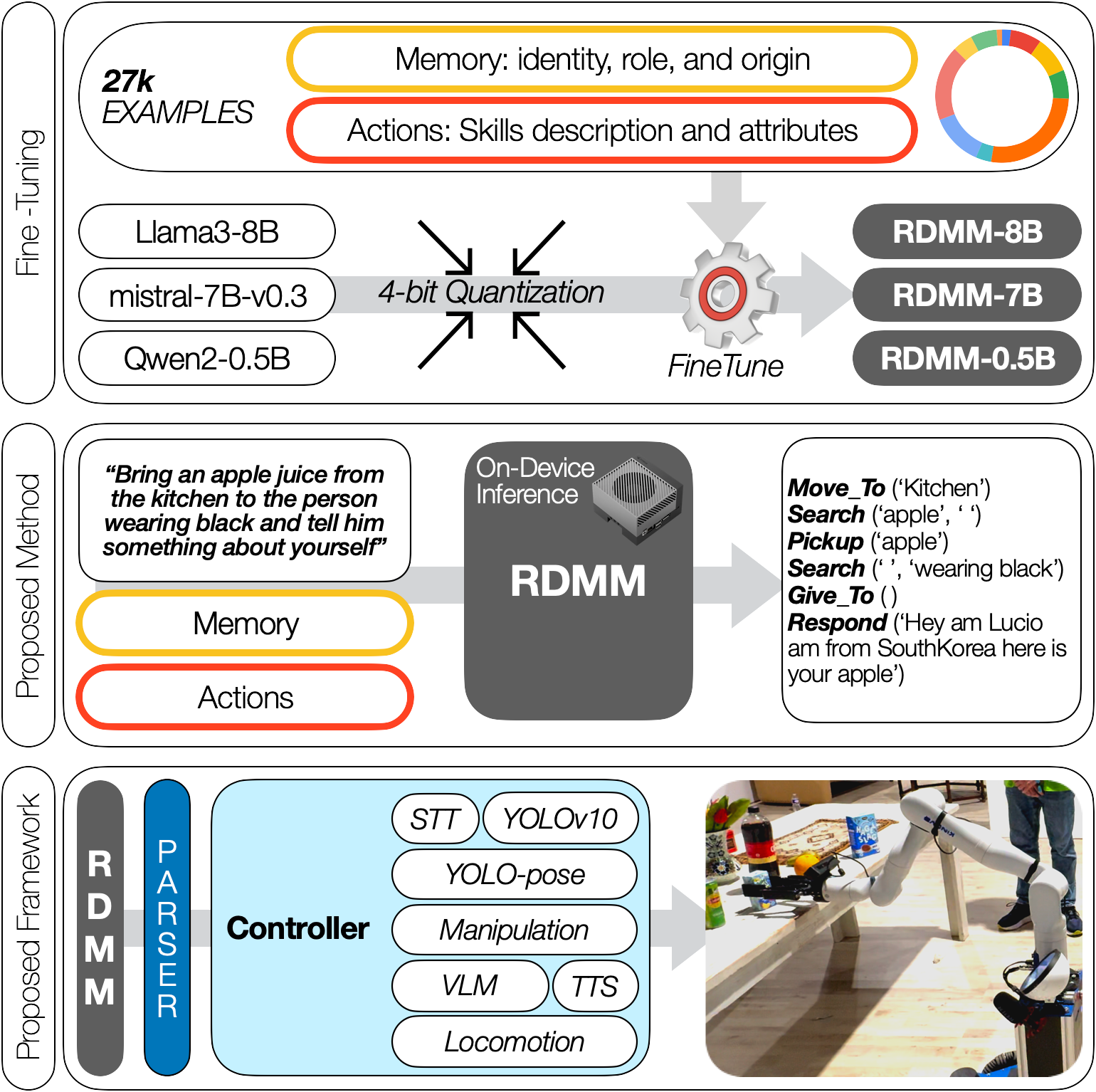}\\
\caption{\textbf{RDMM Overview}: The process begins by fine-tuning quantized LLM models on our specialized dataset to create RDMM models. The illustration showcases an example of RDMM's On-Device inference, followed by the proposed framework parsing the RDMM-generated plans for execution. These plans are carried out using a controller that interacts with various models and enabling both robotic manipulation and locomotion.}
\label{fig:thumbnail}
\vspace{-0.5cm}
\end{figure}

In the rapidly advancing field of robotics and artificial intelligence, the imperative to augment the decision-making capabilities of autonomous systems has been a paramount concern. These models can enhance decision-making, interaction, and planning through their linguistic and contextual understanding abilities. Nevertheless, the direct deployment of large language models in domain-specific robotic tasks faces significant challenges. These key challenges include first, insufficient ability to integrate and leverage personal contextual knowledge about the agent itself, such as its background, capabilities, and specific skills. Second, deployment in real-time on-device settings necessitates efficient inference mechanisms, which can be limited by the computational complexity of large language models.

Recently, there are many methods for solving the grounding problems of LLMs in robotics. PaLM-E \cite{PaLM-E} generates control sentences according to multi-modal data. RT-X \cite{rt-x} directly infer instructions based on languages and images. ChatGPT for Robotics \cite{ChatGPTforRobotics} needs the declaration of APIs for reasoning the actions of tasks. SayCan \cite{SayCan} selects most suitable actions according to environmental information. VoxPoser \cite{VoxPoser} converts the observation space into a 3D value maps for generating trajectories. While existing methods can achieve domain-specific planning and handle some partial disturbances, a key limitation is their inability to incorporate the agent's own knowledge, such as personal background information, capabilities, and skills. This personal contextual knowledge is crucial for well-reasoned question-answering to support effective planning processes.

For instance, a domestic robot assistant could be given a simple task such as delivering an apple to the individual wearing a black t-shirt, and then engaging in a conversation about its recent achievements or favorite color. Existing methods would face difficulties in executing this request, as the employed large language models lack access to the robot's personal knowledge. In contrast, our RDMM framework enables the agent to retrieve and utilize its own information, including its identity, role, and origin, to formulate an appropriate and informative response. This could involve statements like \textit{'I am Lucio, a household robot assistant. How may I assist you?'} or \textit{'Hello, I am Lucio, and I originate from South Korea'}.
Furthermore, a straightforward task that would challenge other methods is \textit{'What can you do?'}, which necessitates the robot's understanding of its own capabilities. Our RDMM framework would provide an informative response highlighting its abilities, such as: \textit{'I can help you with tasks such as moving to a location, searching for objects or people, picking up objects, placing them on a surface, and answering questions.'}.

This paper focuses on developing RDMM models by fine-tuning large language models to acquire self-aware domain-specific planning capabilities. First, the study constructed a comprehensive dataset centered on the tasks and rules of the RoboCup@Home competition. Building upon this foundation, the dataset was further expanded to incorporate the agents' personal knowledge and information regarding their own capabilities and skills. This approach empowers the large language models to not only plan effectively for the given tasks, but also engage in meaningful interactions by providing insightful responses to inquiries about their personal details and abilities, such as their identity, role and background.

This paper makes the following key contributions:
\begin{itemize}
\item A local framework that leverages RDMM models to enhance the autonomous decision-making capabilities of robots, integrating knowledge of their skills and personal information.
\item Comparative analysis of our method against base language models, GPT-4o-mini and GPT-4o and other LLM-based approaches, showcasing the advantages of RDMM models in terms of planning accuracy, On-Device compatibility and inference speed.
\item Real-world evaluation of our system at the RoboCup@Home competition, demonstrating its ability to handle complex robotic tasks within a household environment.
\item Open-source framework, benchmarks, RDMM models, a specific-domain planning dataset of 27k text pairs, and a dataset of annotated 1.3k images to facilitate further research and development in this area.
\end{itemize}

\section{Related Work}
\label{sec:related_work}

\begin{table}
    \setlength{\tabcolsep}{4pt} 
    \vspace{0.2cm}
    \centering
    \caption{LLM-Based Methods for complex robotics tasks comparison}
    \vspace{-0.2cm}
    \begin{tabular}{p{2cm} c c l}
        \toprule
        \textbf{\textsc{Methods}}
        & 
        \begin{tabular}{@{}c@{}}
            \textsc{\textbf{Inputs}} \\
            \textsc{Text +}
        \end{tabular}
        & 
        \textbf{\textsc{Output}}
        &
        \begin{tabular}{@{}c@{}}
            \textsc{\textbf{Model Info.}} \\
            \textsc{(on-device)}
        \end{tabular}
        \\
        
        \midrule
        
        LLM-BT\cite{LLM-BT}                             & Images                  & Variable BTs          & (x) ChatGPT                    \\ \midrule
        SayCan\cite{SayCan}                             & Images                  & Actions               & (x) PaLM                       \\ \midrule
        VoxPoser\cite{VoxPoser}                         & Images                  & Trajectories          & (x) GPT-4                      \\ \midrule
        PaLM-E\cite{PaLM-E}                             & Multi-modal             & Description           & (x) PaLM(540B)                 \\ \midrule
        Huang et al.\cite{Huang}                        & --                      & Actions               & (x) GPT-3(175B)                \\ \midrule
        Raman et al.\cite{Raman}                        & --                      & Actions               & (x) GPT-3 family               \\ \midrule
        Text2Motion\cite{Text2Motion}                   & Scene desc.             & Actions               & (x) GPT-3.5 family             \\ \midrule
        ProgPrompt\cite{ProgPrompt}                     & --                      & Code                  & (x) GPT-3                      \\ \midrule
        LM-Nav\cite{LM-Nav}                             & Image                   &                       & (x) GPT3                       \\ \midrule
        TidyBot\cite{tidybot}                           &                         &                       & (x) GPT3                       \\ \midrule
        RT-X2\cite{rt-x}                                &                         &                       & (x) RT2X-55B                   \\ \midrule
        LLM+P\cite{LLM+P}                               & Scene desc.             & Description           & (x) GPT-4                      \\ \midrule
        ViLaIn\cite{ViLaIn}                             & Image                   & Description           & (x) ChatGPT4                   \\ \midrule
        Code as Policies\cite{CodeAsPolicies}
                                                        & Images                  & Code                  & (x) GPT-3                      \\ \midrule
        ChatGPT for Robotics\cite{ChatGPTforRobotics}
                                                        & APIs                    & Actions               & (x) ChatGPT4                   \\ \midrule
        RDMM(Ours)                                      & Actions                 & Actions               &(\checkmark) RDMM-8B            \\
                                                        & Memory                  &                       &(\checkmark) RDMM-7B            \\
                                                        &                         &                       &(\checkmark) RDMM-0.5B            \\ \midrule
            
    \end{tabular}
    \label{tab:other_methods}
    \vspace{-0.8cm}
\end{table}

Large language models represent a significant advancement in integrating physical robots with AI systems. This approach aims to address the limitations of large language models, which often lack the necessary contextual grounding for effective decision-making in real-world environments. By conditioning language models with pre-trained behaviors, LLM-based systems enable robots to engage in more natural interactions, understand task-specific constraints, and generate executable plans tailored to their capabilities.

The field of LLM-based robotics has witnessed the development of several notable approaches that demonstrate the potential of integrating large language models with robotic systems \cite{LLM-BT, SayCan, VoxPoser, PaLM-E, Huang, Raman, Text2Motion, ProgPrompt, LM-Nav, tidybot, rt-x, CodeAsPolicies, LLM+P, ViLaIn, ChatGPTforRobotics}. For instance, LM-Nav\cite{LM-Nav} proposes a goal-conditioned policy that utilizes large, un-annotated datasets, combining pre-trained models for navigation, image-language association, and language modeling. This enables robots to navigate complex environments based on natural language instructions without the need for expensive supervision or fine-tuning, showcasing the practical applications of pre-trained models. Similarly, TidyBot\cite{tidybot} focuses on personalizing robotic assistance for household tasks by learning user preferences through language-based planning and perception, leveraging the few-shot summarization capabilities of LLMs to quickly adapt to new scenarios. Furthermore, LLaRP\cite{LlaRP} adapts large language models for reinforcement learning in robotics tasks, utilizing a frozen LLM to take text instructions and visual observations, and outputting actions directly in the environment. This system demonstrates robustness in diverse rearrangement tasks, highlighting the potential of LLMs in reinforcement learning for robotics. Additionally, the Code as Policies\cite{CodeAsPolicies} approach leverages LLMs trained on code-completion to generate robot policy code from natural language commands, enabling the synthesis of policy code that processes perception outputs and parameterized control primitives, showcasing the expressive power of LLMs in translating high-level instructions into executable robot behaviors. 
Despite advancements, robots still need to improve natural interactions by better leveraging their knowledge and capabilities. Efficient inference requires local operation for speed and affordability. As shown in Table \ref{tab:other_methods}, most previous methods depend on large models with server-based inference, increasing costs. Our approach eliminates the need for cloud services by running smaller models to run directly on the robot, resulting in reduced latency, improved autonomy, improved privacy and security, and greater reliability for practical applications.

\section{Method}
\label{sec:Method}
\subsection{Dataset Creation}
To create a comprehensive dataset for household robots, we drew inspiration from the RoboCup@Home competition tasks, ensuring it covers a wide range of essential skills needed for domestic activities. The dataset was designed into three categories: 
action-oriented tasks and self-awareness-oriented tasks, each essential for enhancing the robot's operational efficiency and decision-making capabilities in real-world environments. The action-oriented section trains the robot to handle tasks like manipulation, navigation, searching, describing, and counting objects, ensuring it can generate effective strategies for these specific robotic tasks. In contrast, the self-awareness-oriented section equips the robot with a deeper understanding of its identity, capabilities, and purpose, enabling it to engage in more human-like interactions, such as guiding, following and meeting individuals. The final category involves tasks that require a combination of action and memory, where the robot must integrate both types of knowledge to execute complex plans, such as delivering an item and engage in a conversation where it require recalling a relevant detail from its memory.

\begin{figure}[H]
\centering
\setlength{\tabcolsep}{0pt} 
\includegraphics[width=0.9\columnwidth]{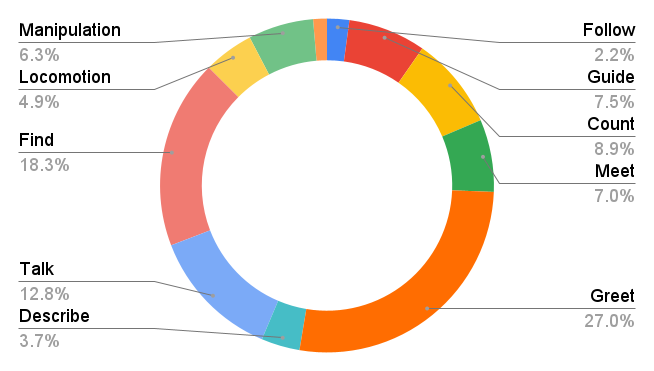}\\
\caption{\textbf{Dataset Distribution by Task:} An overview of the dataset allocation, illustrating the ratio of data dedicated to each specific task. Ensuring balanced and comprehensive training for task-specific model performance.}
\label{fig:dataset_ratio}
\end{figure}

The dataset comprises 27,514 manually annotated examples, each consisting of textual input-output pairs specifically focused on household tasks. Dataset are structured into 42 scenario-based segments, with each scenario categorized under distinct task types, shown in Fig.\ref{fig:dataset_ratio}. The dataset encompasses 21 distinct skills, each outlined with detailed attributes in Table \ref{tab:actions_desc}. To enhance the robot's decision-making and operational efficiency, system messages provide action descriptions, usage information, and access to the robot’s personal memory, allowing it to recall its knowledge in efficiently. This dataset not only serves as a benchmark for evaluating our models but also plays a crucial role in training the robot for real-world applications. By incorporating both action-based and memory-based tasks, the dataset helps the robot develop a deeper understanding of its role, fostering more rational, context-aware decision-making.


\begin{table} [H]
    \vspace{0.2cm}
    \centering
    \caption{Summary of Dataset Actions}
    \vspace{-0.2cm}
    \begin{tabular}{p{3.5cm} l}
        \toprule
        \textsc{\textbf{Actions}}           & \textsc{\textbf{Description}} \\ \midrule
        Respond(request)                    &  Respond to user \\
        Move\_To(location)                  &  Move to a location \\
        Pour\_In(object)                    &  Pour object into a container \\
        Search\_Object(name$^o$, desc.$^*$) &  Search for an object \\
        Search\_Person(name$^o$, desc.$^*$) &  Search for a person \\
        Pickup()                            &  Pickup an object \\
        Place\_On(placement)                &  Place picked up object on placement \\
        Place\_Next(object)                 &  Place picked up object next to object \\
        Give\_To()                          &  Give an object to user \\
        Open(object)                        &  Open a door \\
        Close(object)                       &  Close a door \\
        Vision\_Ask(Question$^*$)           &  Ask VLM and return in Answer()\\
        Answer()                            &  Retrieve answer\\
        Follow()                            &  Follow a person \\
        New\_Request()                      &  Take a new request \\
        Count\_Person(desc.$^*$)            &  Count people and return in Answer()\\
        Count\_Object(name$^o$, desc.$^*$)  &  Count object and return in Answer()\\
        Ask\_Name()                         &  Ask name and return in Answer() \\
        What\_Time()                        &  Retrieve time \\
        What\_Day()                         &  Retrieve date \\
        What\_Tomorrow()                    &  Retrieve tomorrow date \\ \bottomrule
    \multicolumn{2}{l}{\tiny $^*$: Arguments is processed using VLM, $^o$: Arguments is processed using YOLO} \\
    \end{tabular}
    \label{tab:actions_desc}
\end{table}

\subsection{Quantization and Fine-Tuning Details} 
Llama3-8B \cite{llama3}, Mistral-7B-v0.3 \cite{mistral}, and Qwen2-0.5B \cite{Qwen2} was selected as base models for fine-tuning due to their optimal balance of size and performance for Jetson Edge devices. To enhance inference efficiency, GPTQ \cite{gptq} method is applied for quantization, which compresses the model to 4-bit precision while preserving performance. We also utilize QLoRA\cite{QLoRA}, freezing the pre-quantized model and train only a new subset of parameters act as an adapter. Training conducted with a learning rate of 2.5e-5 and capped at 1000 steps, while targeting specific layers such as \textit{q\_proj, k\_proj, v\_proj, o\_proj, gate\_proj, up\_proj} and \textit{down\_proj}. QLoRA combines the 4-bit NormalFloat quantization, Double Quantization, and Low-Rank Adapters (LoRA)\cite{LoRA} to achieve efficient 4-bit quantization. For a single linear layer in the quantized base model with a single LoRA adapter, QLoRA is defined as:

\begin{equation}
    \begin{aligned}
        \textbf{Y}^{BF16} = \textbf{X}^{BF16} \ast doubleDeq(c_{1}^{FP32},c_{2}^{k-bit},\textbf{W}^{NF4}) \\
        +\textbf{X}^{BF16}\textbf{L}_{1}^{BF16}\textbf{L}_{2}^{BF16}
    \end{aligned}
\end{equation}

where $doubleDeq$ is the double de-quantization process:

\begin{equation}
    \begin{aligned}
        &doubleDeq(c_{1}^{FP32},c_{2}^{k-bit},\textbf{W}^{k-bit}) \\
        &= dequant(dequant(c_{1}^{FP32},c_{2}^{k-bit}),\textbf{W}^{4-bit})\\
        &= \textbf{W}^{FB16}
    \end{aligned}
\end{equation}

QLoRA uses NF4 for the weights (\textbf{W}) and FP8 for the quantization constants ($c_2$). The block-size is set to 64 for \textbf{W} for higher precision and 256 for $c_2$ to conserve memory. 
During the backward pass, only the gradients with respect to the LoRA adapter weights ($\frac{\delta E}{\delta \textbf{L}_i}$) are computed, not for the 4-bit weights ($\frac{\delta E}{\delta \textbf{W}}$). 
However, computing ($\frac{\delta E}{\delta \textbf{L}_i}$) involves calculating $\frac{\delta \textbf{X}}{\delta \textbf{W}}$, which requires dequantizing the storage $\textbf{W}^{NF_4}$ to the computation data type $\textbf{W}^{BF16}$. 
In summary, QLoRA uses 4-bit NormalFloat as the storage data type and 16-bit BrainFloat as the computation data type. The storage data type is dequantized to the computation data type for the forward and backward passes, but gradients are only computed for the LoRA parameters in 16-bit precision.
Training time were 24 minutes for RDMM-8B, 11 minutes for RDMM-7B, and 5 minutes for RDMM-0.5B on a single NVIDIA RTX 4090 GPU.


\begin{figure*}
\centering
\includegraphics[width=7in]{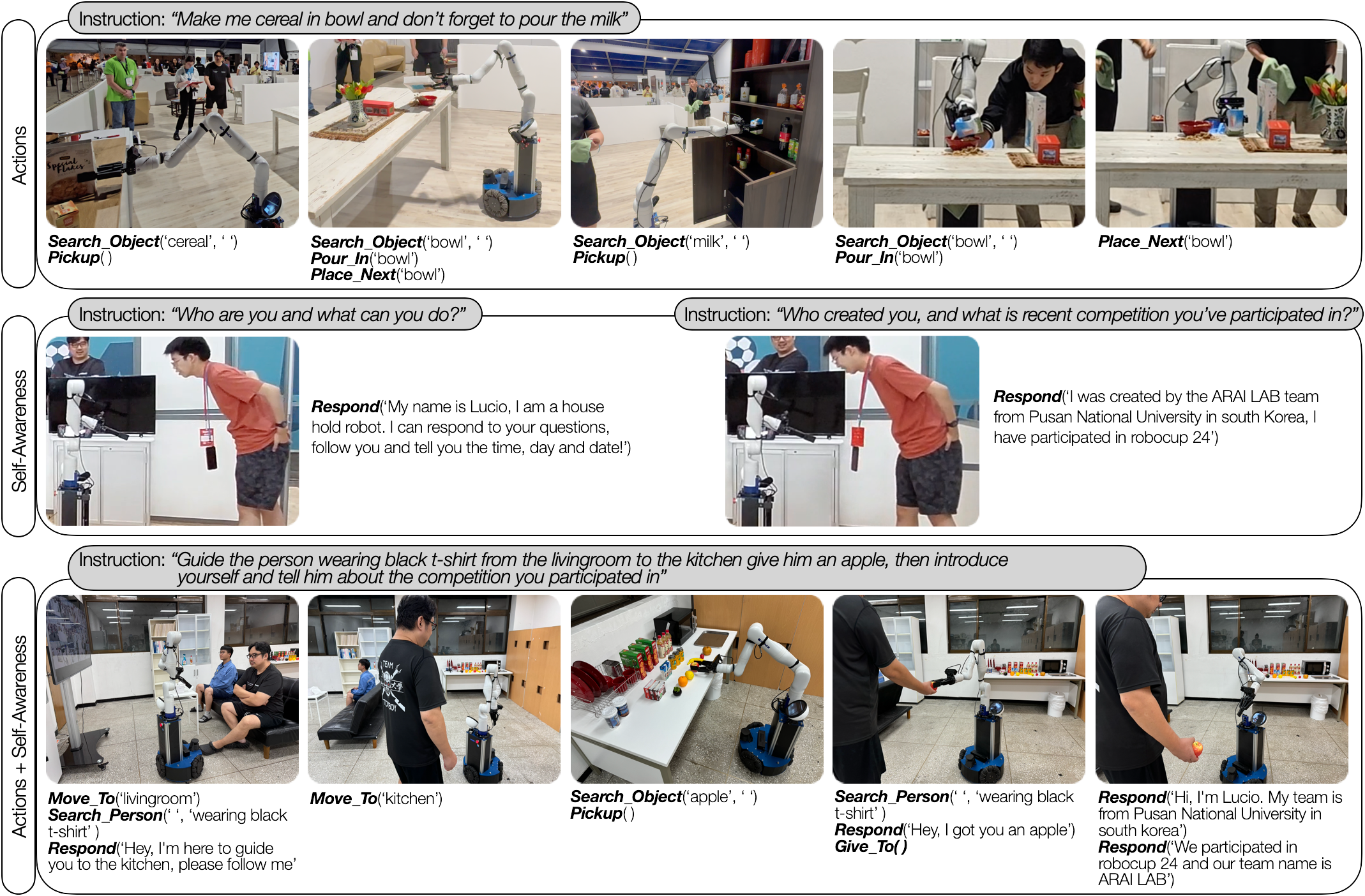}
\caption{\textbf{Household Robot Planning with RDMM:} This illustration shows three examples of Lucio, a home service robot, using local RDMM model inference to plan and execute tasks. These include planning actions to make cereal, answering self-awareness questions about Lucio’s personal memory, and combining actions with self-awareness by retrieving an apple for a person and engaging in conversation about itself.}
\label{fig:lucio_grid}
\vspace{0.2cm}
\end{figure*}

\subsection{Framework Overview}
\subsubsection{Parser \& Controller}
The parser component of our framework is responsible for translating the RDMM-generated plans into actionable commands that the robot can execute. The controller then interprets these commands and interacts with various models, such as VLMs, YOLO, STT and TTS models, to perform specific tasks.

\subsubsection{Vision Language Model}
Visual perception models are crucial for enabling robots to navigate and interact with their surroundings effectively. We employ a 4-bit quantized internlm-xcomposer2-vl-7b \cite{VLM:intern} Vision-Language Model (VLM) to interpret contextual cues and extract detailed visual information. This model provides accurate descriptions of people, objects, and scenes, making it a reliable source of visual intelligence. For example, the VLM can accurately identify if a person is wearing shoes or holding a cup. In Fig. \ref{fig:lucio_grid}, within the actions + self-awareness example, the generated plan includes the action \textit{Search\_Person(' ', 'wearing black t-shirt')}, where the second argument is processed by the VLM to interpret the person's description.


\subsubsection{YOLO Model}
For our real-time object detection algorithms supporting robotic manipulation tasks, the first priority is accurately identifying objects in the environment. To achieve this, we trained a YOLOv10L model on an annotated dataset containing 1.3k images sourced from the RoboCup@Home competition. In Fig. \ref{fig:lucio_grid}, within the actions example, the generated plan includes the action \textit{Search\_Object('cereal', ' ')}, where the first argument is processed by YOLO to detect object location. Additionally, for human detection and pose estimation, we utilize the YOLOv8-pose model.

\subsubsection{Automatic Speech Recognition}
We use Whisper for speech recognition, transcribing audio into text and providing feedback to indicate the robot is listening. For natural responses, we use Seliro-TTS for human-like text-to-speech.

\begin{figure*}
\centering
\includegraphics[width=\textwidth]{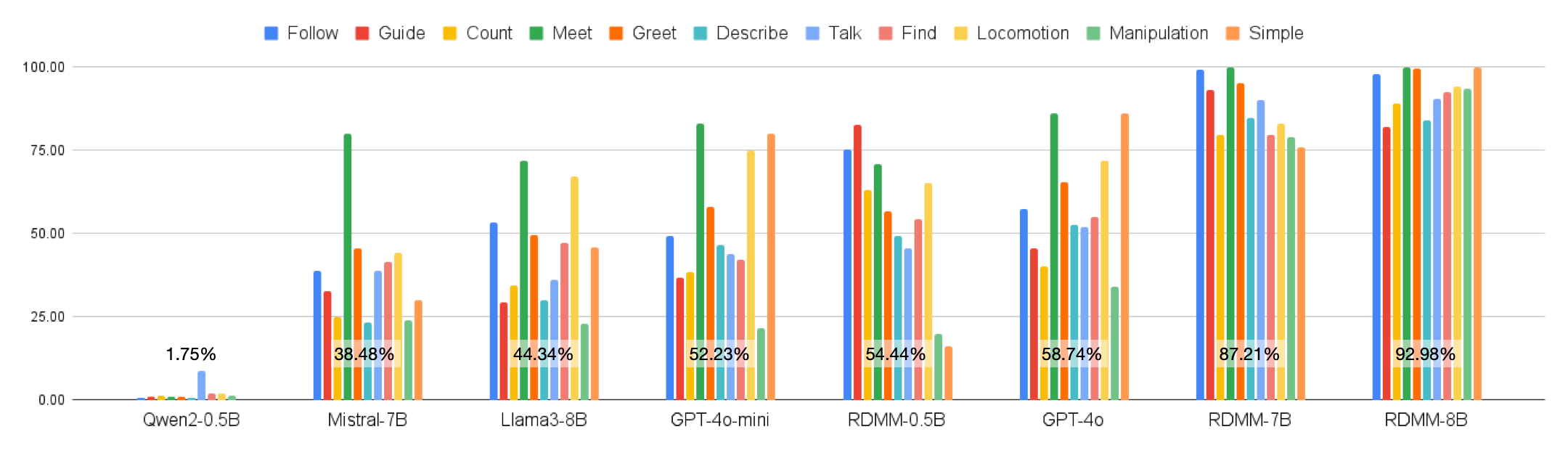}
\caption{\textbf{Benchmark Accuracy Across Tasks:} This graph presents the evaluation results for RDMM-8B, RDMM-7B, and RDMM-0.5B models, compared with 20-shot conditioned baseline models Llama3-8B, Mistral-7B, and Qwen2-0.5B, alongside GPT-4o and GPT-4o-mini. It highlights their accuracy across various tasks, offering insights into each model's performance in different task scenarios.}
\label{fig:models_acc_graph}
\end{figure*}

\section{Experiments}
\label{sec:Experiments}
We evaluated the accuracy, on-device compatibility and inference speed of our RDMM models, comparing them to baseline models, GPT-4o-mini and GPT-4o. Additionally, we tested our model's real-world performance during the RoboCup@Home competition.


\subsection{Models Planning Accuracy}
The accuracy comparison graph in Fig.\ref{fig:models_acc_graph} compares the accuracy of several models across various tasks. It highlights the strong performance of the RDMM models (RDMM-8B, RDMM-7B, and RDMM-0.5B), with a particular focus on their improvements over base models and GPT-4o-mini and GPT-4o. both baseline and GPT models were conditioned with 20-shots examples from the dataset to ensure a fair evaluation across each task. 
The RDMM-8B model achieves the highest accuracy, with an average of 92.98\%, showcasing a significant improvement from its base model's 44.34\%. This indicates a substantial leap in capabilities, particularly in tasks like "Follow," "Meet," and "Simple." Similarly, the RDMM-7B model reaches an impressive 87.21\% accuracy, surpassing both its base model’s performance (38.48\%) and other comparative models, such as GPT-4o. The RDMM-0.5B model, while smaller in scale, still demonstrates a marked improvement over its base model, increasing accuracy from 1.75\% to 54.44\%. Although it slightly trails behind GPT-4o, which achieved 58.74\%, it still outperforms GPT-4o-mini at 52.23\%, indicating the model's competitive edge despite its smaller size.

\subsection{On-Device Inference Compatibility}
The compatibility of RDMM models for on-device inference was evaluated across various Jetson hardware platforms, including the Orin AGX 64GB, Xavier AGX 32GB, Xavier AGX 16GB, Orin NX 16GB, and Xavier NX 8GB, all of which employ ARM architecture with integrated RAM and VRAM.

\subsubsection{RDMM On-Device Compatibility}
The RDMM models—RDMM-8B, RDMM-7B, and RDMM-0.5B—were tested to ensure local inference on these devices. RDMM-8B, requiring 1.1GB RAM and 8.5GB VRAM, and RDMM-7B, requiring 1GB RAM and 6.8GB VRAM, successfully operated on most platforms. However, the Xavier NX 8GB, with limited memory, could only support the RDMM-0.5B model, which demands 0.34GB RAM and 1.9GB VRAM. The larger RDMM models exceeded the available memory on the Xavier NX 8GB, highlighting the importance of aligning model size with hardware constraints for effective on-device inference.

\subsubsection{Framework On-Device Compatibility}
\begin{figure}[H]
\centering
\setlength{\tabcolsep}{0pt} 
\includegraphics[width=\columnwidth]{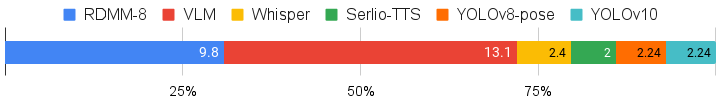}\\
\caption{\textbf{Framework VRAM consumption:} A graphical representation depicting the VRAM usage of each model within the framework.}
\label{fig:vram}
\end{figure}

We also evaluated the full system framework, including VLM, Whisper, Serlio-TTS, YOLOv8-pose, and YOLOv10, alongside the RDMM model. The results, illustrated in Fig.\ref{fig:vram}, shows the memory usage ratios of each model on a local device. The entire system required 30GB of memory, making the 32GB Xavier AGX the smallest device capable of running it.

\subsection{Models Inference Speed Comparison}







\begin{figure}
\centering
\setlength{\tabcolsep}{0pt} 
\includegraphics[width=\columnwidth]{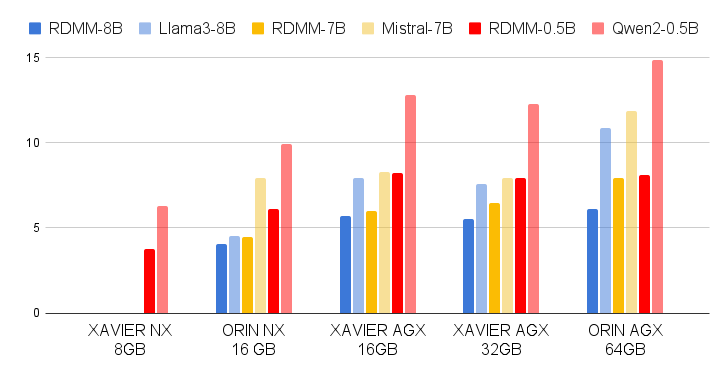}\\
\caption{\textbf{On-Device inference speed comparison:} A detailed analysis comparing the inference speeds of RDMM and baseline models across various Jetson devices. This comparison highlights the efficiency and performance of each model when deployed directly on hardware.}
\label{fig:models_speed_graph}
\end{figure}

The performance evaluation graph presented in Fig.\ref{fig:models_speed_graph} demonstrates the inference speed comparison of RDMM models against other models on various Jetson devices, highlights a slight trade-off between speed and enhanced capabilities. While RDMM models are marginally slower than their base models—such as Llama3-8B, Mistral-7B, and Qwen2-0.5B this slowdown is primarily due to the Progressive Fine-Tuning with Layer-wise Re-calibration approach, which integrates a QLoRA compact neural network adapter. For instance, on the ORIN AGX 64GB, the RDMM-8B model achieved 6.12 tokens per second (T/s), compared to Llama3-8B’s 10.86 T/s and Mistral-7B’s 11.87 T/s. Similarly, on the XAVIER AGX 32GB, the RDMM-8B model achieved 5.54 T/s, compared to Llama3-8B’s 7.56 T/s and Mistral-7B’s 7.95 T/s.
On smaller on-device platforms like the XAVIER AGX 16GB and ORIN NX 16GB, RDMM models still showed competitive results. For instance, on the ORIN NX 16GB, RDMM-0.5B delivered 6.12 T/s compared to Qwen2-0.5B's 9.90 T/s. Even on the entry-level XAVIER NX 8GB, where only RDMM-0.5B could run, it managed 3.75 T/s, showcasing the model's inference on limited hardware.

\subsection{Real World Evaluation}
The real-world evaluation of the RDMM models took place during the RoboCup@Home Competition, using Lucio, a custom-built home service robot platform. In this environment, the RDMM models were responsible for handling various household and service-oriented tasks that required not only decision-making but also a level of self-awareness. These tasks involved navigating through complex environments, following people while carrying luggage, and guiding individuals to specific locations. Lucio's ability to understand its role was essential in tasks such as acting as a receptionist or handing items to people, where it needed to interact naturally and engage in small talk, as shown in Fig.\ref{fig:lucio_grid}. An example of this is guiding a person while engaging in small talk about a specific topic, highlighting how self-awareness improves interaction and enhances service quality in real-world situations.

\section{Conclusion}
\label{conclusion}
This research presents the development and deployment of RDMM models, addressing key challenges that LLMs face when applied to domain-specific tasks. By integrating personal contextual knowledge into the decision-making process, RDMM models offer enhanced capabilities for self-aware planning, interaction, and task execution. Unlike existing methods, which struggle to incorporate an agent's personal background and specific skills
Our approach demonstrates the viability of running powerful language models locally on edge devices without compromising accuracy at a promising inference speed, even on devices with as little as 8GB of memory. This achievement not only enhances the autonomy of robots in practical applications but also reduces reliance on external cloud-based systems, making it an affordable solution. The comprehensive dataset we constructed, including task-specific scenarios and self-awareness-oriented examples, lays the groundwork for future advancements in self-aware robotic planning and interaction.

\bibliographystyle{IEEEtran}

\bibliography{reference}

\end{document}